\pdfoutput=1
\documentclass[11pt]{article}

\usepackage[preprint]{acl}
\usepackage{amsmath}
\usepackage{times}
\usepackage{latexsym}
\usepackage{booktabs}
\usepackage{multirow}
\usepackage{hyperref}
\usepackage{caption}
\usepackage{graphicx}
\usepackage{listings}

\usepackage[T1]{fontenc}
\usepackage[utf8]{inputenc}

\usepackage{microtype}

\usepackage{inconsolata}

\usepackage{graphicx}

\definecolor{codegreen}{rgb}{0,0.6,0}
\definecolor{codegray}{rgb}{0.5,0.5,0.5}
\definecolor{codepurple}{rgb}{0.58,0,0.82}
\definecolor{backcolour}{rgb}{0.95,0.95,0.92}

\title{ADO: Automatic Data Optimization for Inputs in LLM Prompts}

\author{% 
  Sam Lin* \\
  Rutgers University \\
  gl550@scarletmail.rutgers.edu \\
  \And
  Wenyue Hua* \\
  University of California, Santa Barbara \\
  wenyuehua@ucsb.edu \\
  \AND
  Lingyao Li \\
  University of Southern Florida \\
  lingyaol@usf.edu \\
  \And
  Zhenting Wang \\
  Rutgers University \\
  zhenting.wang@rutgers.edu \\
  \And
  Yongfeng Zhang \\
  Rutgers University \\
  yongfeng.zhang@rutgers.edu \\
}

\begin{document}
\maketitle
\begin{abstract}
This study explores a novel approach to enhance the performance of Large Language Models (LLMs) through the optimization of input data within prompts. While previous research has primarily focused on refining instruction components and augmenting input data with in-context examples, our work investigates the potential benefits of optimizing the input data itself. We introduce a two-pronged strategy for input data optimization: content engineering and structural reformulation. Content engineering involves imputing missing values, removing irrelevant attributes, and enriching profiles by generating additional information inferred from existing attributes. Subsequent to content engineering, structural reformulation is applied to optimize the presentation of the modified content to LLMs, given their sensitivity to input format. Our findings suggest that these optimizations can significantly improve the performance of LLMs in various tasks, offering a promising avenue for future research in prompt engineering. The source code is available at \url{https://anonymous.4open.science/r/ADO-6BC5/}.

\end{abstract}

% ``GPT-4 + GPT-4''

\begin{figure*}[t]
  \centering
  % \hspace{-50pt}
  \includegraphics[width=0.9\linewidth]{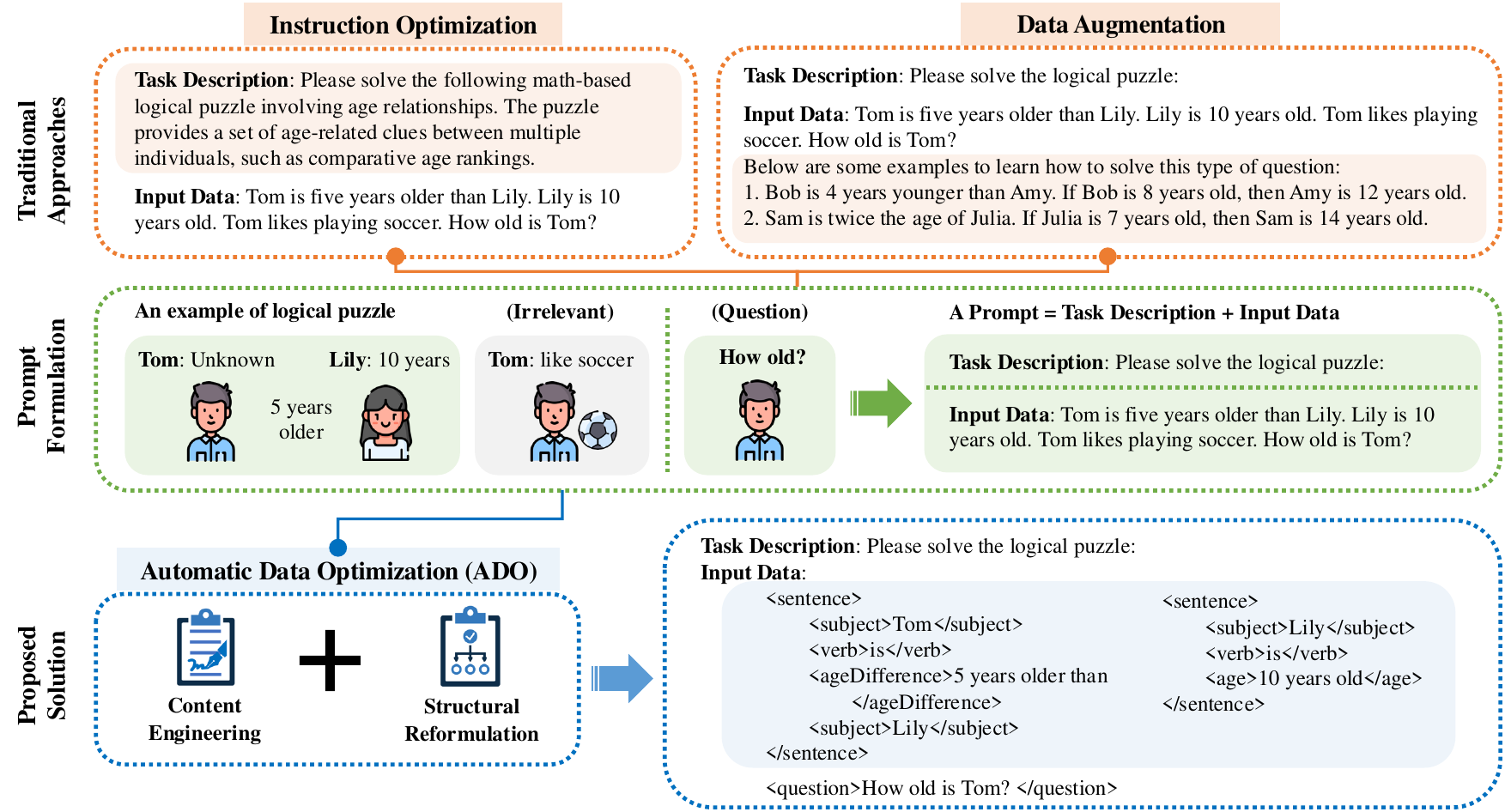}
  \vspace{-5pt}
  \caption{Types of prompt engineering approaches. Given an inference task, such as solving a logical puzzle (as shown in the middle of the figure), prior works primarily focus on either optimizing instructions or augmenting the input data with similar examples, as depicted at the top of the figure. In contrast, we propose optimizing the input data to enhance its presentation to LLMs for more effective task inference, as illustrated at the bottom of the figure.}
  \label{fig:types-of-prompt-engineering-works}
  \vspace{-8pt}
\end{figure*}

\section{Introduction}
Large Language Models (LLMs) \cite{achiam2023gpt, team2023gemini, touvron2023llama} have demonstrated exceptional proficiency across a wide array of tasks. They have been successfully implemented in various real-world applications, including personalized recommendations \cite{wu2024survey, hua2023tutorial}, healthcare \cite{yu2024aipatient, yu2024large, li2024scoping}, financial decision-making \cite{li2023large, wu2023bloomberggpt}, and advanced language reasoning \cite{huang2022towards, fan2023nphardeval, sharan2023llm}. In particular, LLM prompting has become a critical research area \cite{chen2023unleashing, chen2024prompt}. This is because LLMs are highly sensitive to input content and format; even slight modifications, such as changes in word order or indentation, can significantly influence their performance \cite{sclar2023quantifying, Fang2024}.

When LLMs are employed for task inferencing, a user prompt (or query) typically comprises two primary components: a task-specific instruction and the input data to be processed according to that instruction. For example,when employing an LLM for Heart Disease classification \cite{baccouche2020ensemble}, the task-specific instruction can be ``analyze the following user's health profile to determine the likelihood of a heart attack'', while the input data can include the individual's health profile, encompassing attributes such as age, medical history, and lifestyle habits. In the context of personalized recommendations, such as for beauty products \cite{geng2022recommendation}, the instruction can be ``generate beauty product recommendations based on the user's recent interaction history with other beauty products'', with the input data consisting of the user's interaction history and a set of candidate beauty products to make recommendations from.

Various prompting methods have been proposed to enhance the inference performance of LLMs. For example, multiple studies have focused on crafting manual prompting strategies \cite{bsharat2023principled, sahoo2024systematic, marvin2023prompt}, such as Chain-of-Thought (CoT) reasoning \cite{wei2022chain}. Additionally, automated methods have been developed to search for optimal instructions tailored to specific tasks \cite{do2024automatic, li2024pap}. For instance, APE \cite{zhou2023large} introduces an iterative Monte Carlo search to refine prompt instructions. Other works focus on providing in-context demonstrations \cite{dong2022survey}, offering examples to guide the model's responses.

Most prior works on prompt engineering have focused on two aspects: (1) optimizing the instruction component of the prompt and (2) augmenting the input data with additional context, such as in-context exemplars, as illustrated on the ``Traditional Approach'' section of Figure \ref{fig:types-of-prompt-engineering-works}. Nevertheless, the role of input data optimization in enhancing LLM performance remains underexplored. 

To address this gap, we investigate whether optimizing the input data portion of the prompt can also enhance performance, as depicted on the ``Proposed Solution'' section of Figure \ref{fig:types-of-prompt-engineering-works}. Towards this goal, we propose a new framework ``\textbf{A}utomatic \textbf{D}ata \textbf{O}ptimization (ADO)'' as well as a new algorithm, ``\textbf{D}iverse \textbf{P}rompt \textbf{S}earch (DPS)''. This framework can optimize input data through two key strategies: content engineering and structural reformulation. First, we apply content engineering to refine input data, such as imputing missing values based on domain knowledge and removing irrelevant attributes that may hinder decision-making. Second, we leverage structural reformulation to modify the format of input data, aiming to optimize data presentation to LLMs. Together, our proposed framework has demonstrated its effectiveness to complement conventional prompting strategies to enhance LLM inference performance.

% Content engineering involves modifying the content of the input data to facilitate task inferencing. For example, in tabular data modeling with LLMs, this may include imputing missing values based on other attributes and domain expertise as well as removing irrelevant attributes that could hinder decision-making. Following content engineering, we apply structural reformulation, which modifies the format of the input data.  Thus, we propose reformulating the modified content to optimize its presentation to LLMs for improved task inference.

% The following sections are organized as below: section \ref{sec:method} introduces the objectives of data optimization, the ADO framework, and a new algorithm, section \ref{sec:imp} provides implementation details, section \ref{sec:exp} presents the experiment result, section \ref{sec:related} presents the relate work and section \ref{sec:conclusion} concludes the paper.

\begin{figure*}[t]
  \centering
  % \hspace{-50pt}
  \includegraphics[width=0.9\linewidth]{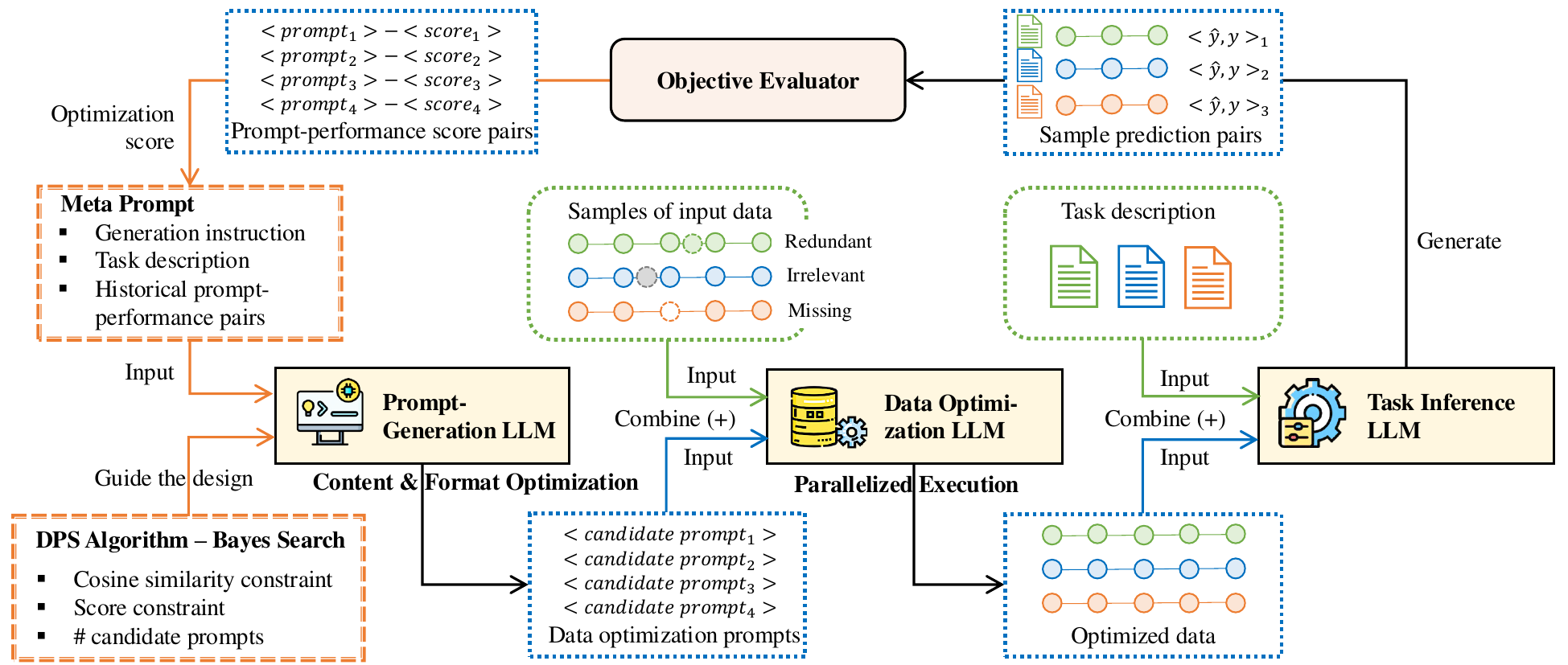}
  \vspace{-5pt}
  \caption{ADO Workflow. The Prompt-Generation LLM initially proposes task-specific instructions for optimizing input data, which the Data Optimization LLM executes on validation set samples, generating optimized inputs. These optimized samples are then processed by the Task Inference LLM to produce task predictions. The Objective Evaluator compares these predictions against the expected outputs (ground truth) using task-specific metrics to compute a score. This score represents the quality of the data optimization instructions, with prior prompt-score pairs provided as additional context to the Prompt-Generation LLM for refining instructions in future iterations.}  
  \label{fig:ADO_workflow}
  \vspace{-8pt}
\end{figure*}

\section{ADO Framework}
\label{sec:method}

This section outlines the objectives of input data optimization and explains the mechanisms by which the ADO framework achieves these objectives.

\subsection{Framework Objective}

In this work, \textbf{we conduct data optimization on the input data part of the prompt} prior to submitting the prompt to a LLM for inference. Our data optimization objectives can be categorized into two aspects: content optimization and format optimization. Content optimization emphasizes enhancing the saliency of features within the data, ensuring that the most relevant and informative attributes are highlighted. Format optimization focuses on structuring the data in an optimal format, such as tables, XML, or other representations that facilitate efficient processing and interpretation. Let $\mathbf{D}$ represents the original input data. The overall data optimization process can be considered as a combination of both content and format optimizations, resulting in an optimized dataset $\mathbf{D}'$:
\begin{equation}
    \mathbf{D}' = f_{format}(f_{content}(\mathbf{D})) = f(\mathbf{D})
\end{equation}
% \zw{maybe provide examples for content optimization and format optimization respectively, or highlight the modified parts belong to "content optimization" and "format optimization" respectively.}
where $f$ is the composite optimization function. This comprehensive approach ensures that the data not only contains salient features but is also presented in a format that maximizes its utility for inference tasks. 

\paragraph{Content Optimization} has been a prominent area of research across various fields and modalities \cite{ahmad2018toward, zhou2004optimization}. For example, in tabular datasets, where each individual is represented by a set of attribute-value pairs, common content optimization procedures include feature extraction, missing value imputation, and attribute aggregation \cite{zheng2018feature}. These techniques aim to enhance the quality of the data by emphasizing salient features and reducing noise. In another example for image inputs, content optimization often involves transformations such as rotation, translation, flipping, cropping, and adjustments to brightness and contrast \cite{jiao2019survey}. These procedures are employed to enhance model performance by augmenting the dataset and improving the representation of important features \cite{barrett2002object, ling2021editgan}.

Traditionally, task-specific data engineering has relied heavily on domain expertise \cite{ling2021editgan}. For example, in the medical field, experts may derive new attributes from existing ones—such as calculating the Body Mass Index (BMI) from weight and height measurements—to create more informative features for analysis. Similarly, for data in natural language form, such as logical puzzles or mathematical problem statements, individuals with linguistic and analytical expertise may augment the text by identifying contextual cues, deducing relevant implicit information, and explicitly defining known and unknown variables to facilitate more effective interpretation. 

However, employing human experts to craft and refine each input data can be both costly and time-consuming. With recent advancements in LLMs, we propose leveraging LLMs as universal domain experts. Specifically, we investigate their ability to propose and execute content optimization procedures across datasets from diverse fields. By automating the content optimization process, we aim to transform the original dataset $\mathbf{D}$ to optimized version $\mathbf{D}$'. The objective is to reduce reliance on human expertise while maintaining or enhancing model performance. This approach not only accelerates the data preparation phase but also has the potential to uncover novel optimization strategies that may be overlooked by human practitioners.

% For each type of modality, we first prompt the LLM\textsubscript{ins} 

\paragraph{Format Optimization} concentrates on the automatic discovery of the optimal format for presenting input data to a LLM, after the content has been optimized. Recent studies have demonstrated that LLMs are highly sensitive to input formatting \cite{sclar2023quantifying}. For example, manipulations such as positional swapping of in-context examples or alphabet shifting have been observed to influence an LLM's performance. Additionally, transforming attribute-value pairs in tabular data into structured formats like XML can enhance LLM performance on classification tasks. Similarly, converting natural language inputs into non-natural language formats using emojis, logical operators, or other symbolic figures has been shown to improve LLM performance \cite{lin2024promptcrypt}. Here, we again leverage LLM to find an optimal formatting function that maximizes the performance. By utilizing LLMs to explore various formatting strategies, we aim to identify structural reformulations that enhance the LLM's performance without altering the underlying content of the data. 

\subsection{Framework Workflow Design}

The ADO framework employs a set of LLMs to automatically optimize the representation of input data \(\mathbf{D}\). As illustrated in Figure \ref{fig:ADO_workflow}, the process initiates with a Prompt Generation LLM, which proposes a data-optimization prompt \(\mathbf{P}_o\) that outlines a set of procedures for modifying \(\mathbf{D}\). Specifically, these procedures consist of two sequential components: the first provides step-by-step instructions for modifying the content of \(\mathbf{D}\), while the second details step-by-step instructions for reformulating the content-optimized data. 

Subsequently, a Data Optimization LLM progressively executes the proposed data-optimization prompt by processing both \(\mathbf{P}_o\) and \(\mathbf{D}\), instructing the model to generate the optimized data \(\mathbf{D}'\) to implement the target function \(\mathbf{D}' = f_{\text{format}}(f_{\text{content}}(\mathbf{D}))\). The optimized data \(\mathbf{D}'\) is then submitted to a Task Inference LLM for processing, and its performance is evaluated on a reserved validation set, serving as the performance measure for \(\mathbf{P}_o\). Finally, \(\mathbf{P}_o\) and its corresponding performance are fed back into the Prompt Generation LLM as additional context, enabling it to generate improved data-optimization prompts in future search rounds.

% The ADO framework employs a set of LLMs to automatically optimize the representation of input data $\mathbf{D}$. The framework operates by searching for a data-optimization prompt $\mathbf{P}_o$ consisting a set of procedures on how to optimally represent $\mathbf{D}$ for task inferencing by LLMs, implementing the target function $f_{format}(f_{content}(\mathbf{D}))$. Specifically, the proposed procedures consist of two sequential parts: the first part provides step-wise instructions on how to modify the content of the input data, while the second part offers step-wise instructions on how to reformulate the content-optimized data. 

% Figure \ref{fig:ADO_workflow} presents the workflow of ADO: 
% guided by $\mathbf{P}_o$, the input data $\mathbf{D}$ is transformed into optimized version $\mathbf{D}'$. To generate the final answer, the LLM performs inference on the optimized data $\mathbf{D}'$ together with a task-specific instruction $\mathbf{t}$.

We now formally define the ADO framework, which involves three instances of LLMs:
\begin{itemize}
\item Prompt Generation LLM (\(\text{LLM}_{\mathcal{G}}\)): Given a meta-prompt $\mathbf{P}_m$ used to instruct generating the data-optimization-prompt \(\mathbf{P}_o\), $\text{LLM}_{\mathcal{G}}$ generates a set of candidate \(\mathbf{P}_o\)s aiming at providing instructions on how to optimize $\mathbf{D}$: 
\begin{equation}
\mathbf{P}_o = \text{LLM}_{\mathcal{G}}(\mathbf{P}_m)
\end{equation}

\item Data Optimization LLM (\(\text{LLM}_{\mathcal{O}}\)): Given a data-optimization prompt \(\mathbf{P}_o\), \(\text{LLM}_{\mathcal{O}}\) optimizes \(\mathbf{D}\) to produce the optimized data \(\mathbf{D}'\):
\begin{equation}
\mathbf{D}' = \text{LLM}_{\mathcal{O}}(\mathbf{P}_o, \mathbf{D})
\end{equation}

\item Task Inference LLM (\(\text{LLM}_{\mathcal{I}}\)): Using the optimized data \(\mathbf{D}'\) and the task-specific instruction \(\mathbf{t}\), \(\text{LLM}_{\mathcal{I}}\) generates the final result \( y \):
\begin{equation}
y = \text{LLM}_{\mathcal{I}}(\mathbf{D}', \mathbf{t})
\end{equation}
\end{itemize}

In the ADO framework, the search for the optimal data-optimization prompt $\mathbf{P}_o$ is typically conducted using a reserved set of data points $S = \{(x, y)\mid x\in\mathbf{D}_S, y\in\mathcal{Y}_{\mathbf{D}_S}\}$ where $\mathcal{Y}_{\mathbf{D}_S}$ is the set of ground truth corresponding to $\mathbf{D}_S$. Given $S$, we sequentially utilize the three LLM instances to generate candidate prompts $\mathbf{P}_o$s, optimize the data $\mathbf{D}$, and produce the final inference result $y'$. By comparing the generated outputs $y$ and with the ground truth labels $y'$, we can evaluate the quality of each candidate $\mathbf{P}_o$ using some task-specific loss function $L(y, y')$. The optimization of $\mathbf{P}_o$ can be formulated as minimizing the loss over $S$:
\begin{multline}
    \mathbf{P}_o^* = \arg\min\limits_{\mathbf{P}_o\in\text{LLM}_{\mathcal{G}}(\mathbf{P_m})}\\
    \sum\limits_{(x_i, y_i)\in S}L(\text{LLM}_{\mathcal{I}}(\text{LLM}_{\mathcal{O}}(\mathbf{P}_o, x_i), \mathbf{t}), y_i)
\end{multline}

Various optimization algorithms such as Automatic Prompt Engineer (APE) \cite{zhou2023large}, Automatic Prompt Optimization (APO) \cite{pryzant2023automatic}, and Optimization by PROmpting (OPRO) \cite{yang2024large, liu2024large, zhou2023large} can be employed to search for a better $\mathbf{P}_o$ based on the loss function $L$. Nevertheless, such algorithms exhibit a potential limitation in optimizing $\mathbf{P}_o$. In the following subsection, we introduce the novel Diverse Prompt Search (DPS) algorithm to address the limitation.

\subsection{DPS Algorithm for $\mathbf{P}_o$ Optimization}

Recently, various optimization algorithms \cite{pryzant2023automatic, yang2024large, liu2024large} have been proposed that leverage LLMs for automatic prompt optimization. Specifically, APE employs an LLM to propose several candidate prompts and selects the one with the best performance based on a reserved validation set. Subsequent works, such as OPRO, build upon this by directly utilizing an LLM as the prompt optimizer. For instance, OPRO instructs an LLM to iteratively propose candidate prompts, one at a time, while providing feedback on the performance of prior proposed prompts on a reserved validation set. This additional context enables the LLM to generate prompts with improved performance in subsequent iterations.

% and evaluate its performance on a validation set, and then instruct the LLM to propose a new prompt that would result in improved performance based on the initial prompt. This process is repeated for an arbitrary number of times, and the prompt with best performance on the validation set will be selected.

Nevertheless, recent studies \cite{zhang2024revisiting, tang2024unleashing} have shown that optimizing by augmenting a single candidate prompt as context in each iteration, without any constraints on the resemblance between candidate prompts, may hinder the discovery of an optimal prompt. Despite being instructed to generate new candidate prompts that differ from previous ones, the LLM may at times converge toward semantically or lexically similar variations of prior proposed prompt(s). In our case, instead of proposing novel data optimization procedures, the LLM may keep proposing procedures that refine the wording or reorder the steps in the prior proposed procedures. This behavior reduces diversity in prompt generation, restricting exploration to a narrow region of the prompt space and yielding only marginal performance improvements.

% In other words, due to the lack of diversity constraints, the iterative generation tends to explore only a limited region of the prompt space, leading to convergence towards a suboptimal solution.

% employs a similar approach in which the prompt-generation LLM$_{\mathcal{G}}$ is instructed to propose a single candidate data-optimization prompt $\mathbf{P}_o$. The performance of this candidate prompt is then evaluated on a reserved validation set $S$. The resulting prompt-score pair is appended to the meta-prompt $\mathbf{P}_m$, serving as a reference for generating new candidate prompts in subsequent iterations. 

% However, initiating the search with only one candidate prompt introduces limitations in the quest for an optimal data-optimization prompt. Despite instructing LLM$_{\mathcal{G}}$ to generate new candidate prompts that differ from those previously generated, the prompts in later iterations tend to converge toward semantically similar variations of the initial prompt. Specifically, instead of proposing novel data optimization procedures, LLM$_{\mathcal{G}}$ often refines the wording or reorders the procedures presented in the initial prompt. This behavior leads to reduced diversity in the prompt generation process and results in marginal performance gains. Due to the lack of diversity constraints, the iterative refinement in OPRO tends to explore only a limited region of the prompt space, leading to convergence towards a suboptimal solution.

To this end, we propose the DPS algorithm, which also employs a LLM as the prompt optimizer, while generating multiple diverse candidate prompts for each iteration of the search process, with both semantic and lexical diversity constraints enforced to grant prompt diversity. Specifically, we request \text{LLM}$_{\mathcal{G}}$ to generate $k$ distinct candidate prompts \{$\mathbf{P}_o^1, ..., \mathbf{P}_o^k\}$ for each iteration of the search. For both semantic and lexical diversity among these prompts, we propose two constraints:

% By ensuring that the cosine similarity between any pair of prompts is below 0.5 and the METEOR score is less than 0.15, DPS encourages exploration across a broader region of the prompt space. This broader search increases the likelihood of identifying more effective data optimization procedures, thereby enhancing the performance of the LLM on the given task.

% In order to search for the best $\mathbf{P}_o$ for a given task, we propose a novel LLM-based optimization algorithm, Diverse Prompt Search (DPS). In DPS, 

\begin{itemize}
    \item Cosine similarity constraint ($c_1$): The cosine similarity between any pair of prompts should be less than $c_1$: $\cos{(\mathbf{P}_o^i, \mathbf{P}_o^j)} < c_1, \; \forall i\neq j$
    \item METEOR Score Constraint ($c_2$): The METEOR score \cite{saadany2021bleu} between any pair of prompts should be less than $c_2$: $\text{METEOR}{(\mathbf{P}_o^i, \mathbf{P}_o^j)} < c_2, \; \forall i\neq j$
\end{itemize}

To dynamically control the extent of prompt diversity tailored to specific tasks, we propose the novel idea of \textbf{incorporating Bayesian Search \cite{turner2021bayesian} to automatically determine optimal values for} \(k\), \(c_1\), and \(c_2\) based on validation set performance. Since Bayesian Search has been widely employed for hyper-parameter tuning in various deep learning models, we propose to integrate this approach with automatic prompt search by treating ADO as a standalone model, with \(k\), \(c_1\), and \(c_2\) as its hyper-parameters. The performance metric for each Bayesian Search iteration is defined as the highest performance achieved among all data-optimization prompts proposed by ADO with a fixed set of hyper-parameters. Such constraints ensure that the generated prompts are semantically and lexically diverse, encouraging exploration of different regions in the prompt space. For Bayesian Search details, please refer to \ref{sec:bayes}.

The generation of qualifying prompts is performed iteratively by repeatedly querying \text{LLM}$_{\mathcal{G}}$ until all $k$ diverse prompts satisfying the above constraints are obtained. Each candidate prompt $\mathbf{P}_o^i$ is evaluated on $S$, based on which result we batch update the generation $\mathbf{P}_o$. The evaluation involves applying the data optimization and inference steps:
\begin{itemize}
    \item Data optimization: $x_i' = \text{LLM}_{\mathcal{O}}(\mathbf{P}_o^i, x_i)$ where $x_i$ is one input data in $S$
    \item Result inference: $y_i' = \text{LLM}_{\mathcal{I}}(x_i', \mathbf{t})$ where $\mathbf{t}$ is the task-specific instruction.
\end{itemize}

The performance of each candidate $\mathbf{P}_o^i$ is assessed by computing a loss function $L$ over $S$:
\begin{equation}
    l_i = \sum\limits_{(x_i, y_i)\in S}L(y_i', y_i)
\end{equation}

The batch of prompt-performance pairs ($\mathbf{P}_o^i$, $l_i$) is then appended to $\mathbf{P}_m$ to guide subsequent iterations of prompt generation. This feedback mechanism informs \text{LLM}$_{\mathcal{G}}$ about the effectiveness of previously generated prompts, enabling it to generate more promising candidates in future iterations.

% \zw{the detailed process of ``refining the set of candidate prompts and incorporating performance feedback with batch update'' may not be clear enough. I suggest to have a latex algorithm here.}
By iteratively refining the set of candidate prompts and incorporating performance feedback with batch update, the DPS algorithm encourages the exploration of a broader search space. This increases the likelihood of discovering more effective data optimization procedures, ultimately enhancing the performance of the LLM on the given task.

\section{Implementation Details}
\label{sec:imp}

This section provides key implementation details of the ADO framework, including the structure of meta-prompts, the execution of parallelized data optimization tasks, the handling of LLM hallucinations through multi-agent debate with cross-validation. By leveraging these components, the ADO framework effectively enhances both the content and format of input data to improve performance across diverse tasks while maintaining factual accuracy and efficiency.

\paragraph{Meta-Prompt} In this purely text-based data optimization framework, the data-optimization prompt $\mathbf{P}_o$ must consist of instructions that can be executed by the LLM without relying on external tools or operations. To ensure this, we incorporate a comprehensive set of modality-specific constraints within the meta-prompt $\mathbf{P}_m$ provided to \text{LLM}$_{\mathcal{G}}$. These constraints guide the prompt generation process, ensuring that \text{LLM}$_{\mathcal{G}}$ avoids proposing optimization procedures that \text{LLM}$_{\mathcal{O}}$ is incapable of performing. For instance, when generating instructions for tabular data, the meta-prompt explicitly prohibits steps such as Principal Component Analysis (PCA), normalization, standardization, or one-hot encoding of categorical attributes, as these require tool-based operations beyond the LLM’s text-based capabilities. Below is an example of $\mathbf{P}_m$:

\paragraph{Parallelized Execution} The generated data-optimization prompt $\mathbf{P}_o$ typically includes multiple procedures, each addressing a specific aspect of data engineering or reformulation (e.g., missing data imputation, structural conversion). We parse the number of procedures generated from $\mathbf{P}_o$ and employ an equivalent number of LLM instances to execute each procedure concurrently. 

Parallel execution provides two advantages: (1) avoiding omission or redundancy -- we observed that prompting \text{LLM}$_{\mathcal{O}}$ to execute a lengthy list of detailed procedures in one go often leads to omissions and repetition. By executing procedures in parallel, we mitigate these issues by breaking down the tasks into smaller, independent units of work for each LLM instance. (2) improving time efficiency -- Sequential execution of a long series of procedures can be time-consuming. Since many procedures are independent of each other and can be directly applied to the raw input data, distributing them across multiple LLM instances significantly reduces the overall time required for data optimization. For procedures that depend on sequential execution -- where the output of one serves as the input for the next -- their execution is grouped together.

\paragraph{Hallucination Mitigation} Instructions included $\mathbf{P}_o$ may sometimes be implemented inaccurately by \text{LLM}$_{\mathcal{O}}$ due to hallucinations. For example, if $\mathbf{P}_o$ includes a directive such as ``Please identify the mathematical terminologies and provide concise definitions, accompanied by examples for each.'' \text{LLM}$_{\mathcal{O}}$ may generate incorrect or inaccurate definitions for some of the terms identified. These inaccuracies could mislead the performance of \text{LLM}$_{\mathcal{I}}$, potentially degrading overall output quality. 

To mitigate the risk of hallucination and improve factual accuracy, we adapt a cross-validation method inspired by \cite{du2023improving}. In this framework, we introduce an additional LLM, denoted as \text{LLM}$_{\mathcal{F}}$ which reviews the optimized input data to identify factual inaccuracies and provides concise explanations for any detected errors. When errors are found, \text{LLM}$_{\mathcal{F}}$'s feedback is passed back to \text{LLM}$_{\mathcal{O}}$, prompting it either to justify its original output or to agree with the corrections suggested by \text{LLM}$_{\mathcal{F}}$. By incorporating this cross-validation framework, we ensure a higher level of factual accuracy, leveraging the complementary strengths of multiple LLMs to reduce the likelihood of hallucinations and errors in the final output.

\begin{table*}[!ht]
\centering
\resizebox{14cm}{!}{
\begin{tabular}{c c c c c c c c c c c c}
\toprule
 LLM for ADO & Algorithm & QA & Job & GSM & AB & AT & AE & CI & HD & FD & Mean\\ 
\hline
\hline
\multirow{4}{*}{GPT-3.5 Turbo} & N/A  & 0.578 & 0.619 & 0.285 & 0.124 & 0.129 & 0.211 & 0.788 & 0.617 & 0.639 & 0.443 \\ 
 & APE & 0.575 & 0.633 & 0.721 & 0.161 & 0.184 & 0.241 & 0.839 & 0.687 & 0.658 & 0.522 \\ 
 & OPRO & 0.583 & 0.627 & 0.734 & \textbf{0.169}& 0.195 & 0.238 & 0.846 & 0.681 & \textbf{0.667}& 0.527 \\ 
  & DPS & \textbf{0.589}& \textbf{0.638}& \textbf{0.755}& 0.166 & \textbf{0.213}& \textbf{0.253}& \textbf{0.853}& \textbf{0.704}& 0.652 & \textbf{0.536}\\ 
\hline
\multirow{4}{*}{Gemini-1.5 Flash} & N/A & 0.569 & 0.607 & 0.299 & 0.137 & 0.115 & 0.197 & 0.791 & 0.625 & 0.612 & 0.439 \\ 
 & APE & 0.581 & 0.621 & 0.698 & 0.159 & 0.176 & 0.219 & 0.827 & 0.701 & 0.661 & 0.516 \\ 
 & OPRO & 0.589 & 0.624 & 0.704 & 0.173 & 0.183 & \textbf{0.238} & \textbf{0.841} & 0.709 & 0.672 & 0.526 \\ 
 & DPS & \textbf{0.595} & \textbf{0.643} & \textbf{0.729} &\textbf{0.198} & \textbf{0.201} & 0.225 & 0.838 & \textbf{0.722} & \textbf{0.699} & \textbf{0.539} \\ 
\hline
\multirow{4}{*}{Llama-3.1 70B} & N/A & 0.563 & 0.588 & 0.281 & 0.117 & 0.135 & 0.188 & 0.769 & 0.629 & 0.615 & 0.431 \\ 
 & APE & 0.571 & 0.613 & 0.675 & 0.129 & 0.166 & 0.205 & 0.798 & 0.673 & 0.649 & 0.498 \\  
 & OPRO & 0.574 & 0.619 & 0.693 & 0.135 & 0.173 & 0.213 & 0.806 & 0.692 & 0.657 & 0.507 \\ 
 & DPS & \textbf{0.581} & \textbf{0.635} & \textbf{0.718} & \textbf{0.159} & \textbf{0.189} & \textbf{0.229} & \textbf{0.827} & \textbf{0.711} &\textbf{0.661} & \textbf{0.523} \\ 
\bottomrule
\end{tabular}
}
\vspace{-3pt}
\caption{ADO performance across all datasets. ``LLM for ADO'' denotes the LLM used within the ADO framework. ``Algorithm'' denotes the algorithm to search for optimal data-optimization procedures. ``Mean'' denotes the mean performance across all datasets. The best performance for each dataset on every LLM is highlighted in bold.}
\label{tab:ado_performance}
\vspace{-9pt}
\end{table*}

\section{Experiments}
\label{sec:exp}

% In this section, we introduce the experiment setting including the 9 benchmarks that we experiment on and the baseline models.

In this section, we aim to evaluate: (1) the effectiveness of ADO as a standalone approach for performance enhancement, (2) whether DPS outperforms existing optimization algorithms in searching for data-optimization procedures, and (3) whether integrating ADO with other prompt engineering methods can further improve performance.

\subsection{Experiment Settings}

\paragraph{Dataset}
To demonstrate the wide applicability of data optimization, we conduct experiments on nine publicly available, real-world datasets across various domains where LLMs are frequently applied \cite{Fang2024, li2023prompt, lin2024data, rouzegar2024enhancing}. These datasets include Big-Bench StrategyQA (QA) \footnote{\url{https://github.com/google/BIG-bench/tree/main/bigbench/benchmark_tasks/strategyqa}}, Fraudulent Job Detection (Job) \footnote{\url{https://www.kaggle.com/datasets/shivamb/real-or-fake-fake-jobposting-prediction}}, Grade School Math 8k (GSM8k) \footnote{\url{https://huggingface.co/datasets/DaertML/gsm8k-jsonl}}, Amazon Beauty (AB) \footnote{\url{https://jmcauley.ucsd.edu/data/amazon/}}, Amazon Toys (AT) \footnote{\url{https://jmcauley.ucsd.edu/data/amazon/}}, Amazon Electronics (AE) \footnote{\url{https://jmcauley.ucsd.edu/data/amazon/}}, Census Income (CI) \footnote{\url{https://archive.ics.uci.edu/dataset/2/adult}}, Heart Disease (HR) \footnote{\url{https://www.kaggle.com/datasets/kamilpytlak/personal-key-indicators-of-heart-disease}}, and Financial Distress (FD) \footnote{\url{https://www.kaggle.com/c/GiveMeSomeCredit/data?select=cs-test.csv}}. For each dataset, we randomly select 1,000 samples to form the validation set $S$. 

% To assess the effectiveness of data optimization across various prompting styles, we evaluate each dataset using (1) zero-shot, (2) few-shot in-context learning, and (3) zero-shot chain-of-thought (CoT) reasoning. For few-shot in-context learning, we randomly select five samples per dataset following the approach in \cite{liu2022few}. For zero-shot CoT reasoning, we apply the method from \cite{wei2022chain}, appending the phrase ``Let's think step-by-step'' to guide the model's reasoning process.

\paragraph{Modeling} The evaluation modeling is twofold. First, we evaluate the effectiveness of ADO under zero-shot prompting, with three LLMs with different backbones for generalizability. To perform data-optimization procedure search, we employ APE, OPRO, and DPS algorithms. Second, we assess whether ADO can be integrated with existing Prompt Engineering techniques (i.e., instruction optimization and data augmentation) to further enhance performance, with GPT-3.5 Turbo as the backbone. For instruction optimization, we employ either Chain-of-Thought reasoning (CoT) \cite{wei2022chain} or PE2 \cite{ye2023prompt} after ADO is applied; similarly, for data augmentation, we employ In-Context Learning (ICL) \cite{liu2022few} subsequent to employing ADO. For CoT, we follow \cite{wei2022chain} by appending the phrase ``Let's think step-by-step'' at the end of the task instruction. For PE2, we employ it to search for the optimal task instruction. For ICL, we randomly select ten samples per dataset and augment them to the prompt for additional context, following the approach from \cite{liu2022few}.

% For prompt search efficiency, we implement an early stopping mechanism. After evaluating every 50 samples, the performance of the current candidate prompt is compared with the best-performing prompt identified so far. If the current prompt underperforms the best prompt for two consecutive evaluations, it is discarded, and the search continues with the next candidate. If the current prompt outperforms the best prompt across the full validation set, it is designated as the new benchmark for subsequent candidates.

% First, we assess the effectiveness of ADO by comparing the performance of optimized data with their unoptimized counterparts, employing three LLMs with different backbones to  generalizability. For data-optimization procedure search, we compare DPS algorithm against APE and OPRO to investigate whether DPS outperforms these approaches. 

\paragraph{Evaluation metrics} We employ accuracy (with balanced accuracy for datasets that have imbalanced binary targets) and Hit@10 for the recommendation datasets from Amazon.

\paragraph{Baselines} To evaluate the effectiveness of ADO, we compare \text{LLM}$_{\mathcal{I}}s'$ performance without data optimization to the performance achieved after ADO is applied. To evaluate the effectiveness of the DPS algorithm on data-optimization procedure search, we compare it against two recent optimization algorithms: APE and OPRO. It is important to highlight that ADO represents a novel sub-direction in the field of prompt engineering and can be combined with existing prompt engineering techniques. Unlike a competitive relationship, ADO and techniques such as CoT, PE2, and ICL are in fact \textbf{complementary}, enabling joint application for enhanced performance. Thus, we utilize CoT, PE2, and ICL as baselines to observe whether combining ADO with any of these techniques achieves better performance compared to using them alone.

% \text{LLM}$_{\mathcal{G}}$, \text{LLM}$_{\mathcal{O}}$, \text{LLM}$_{\mathcal{I}}$

\paragraph{LLM Backbones}
We employ three instances of the same LLM as \text{LLM}$_{\mathcal{G}}$, \text{LLM}$_{\mathcal{O}}$, and \text{LLM}$_{\mathcal{I}}$. For generalizability, we test with three different LLMs, including GPT-3.5 Turbo, Gemini-1.5 Flash, and Llama-3.1 70B. Additionally, Gemini-1.5 Pro is instantiated as \text{LLM}$_{\mathcal{F}}$, which will be employed in Section \ref{sec:Ablation Study}. We set the temperature to 1.0 for \text{LLM}$_{\mathcal{G}}$ to encourage the generation of more creative content. For \text{LLM}$_{\mathcal{O}}$ and \text{LLM}$_{\mathcal{I}}$, we set the temperature to 0 to obtain more consistent outputs.

% \paragraph{Prompt Optimization Algorithms}
% To facilitate the task of input data optimization, we randomly select 1,000 samples from each dataset to form the validation set. We then apply three distinct LLM-based prompt optimization algorithms to automatically search for the task-specific data-optimization prompt that achieves the best performance on the reserved validation set. Specifically, we employ our novel DPS algorithm and compare its performance against two existing approaches: APE \cite{} and OPRO \cite{yang2024large}. To ensure a fair comparison, we maintain an equal total number of candidate prompts across all algorithms.

% For the APE algorithm, 5 candidate prompts are generated per iteration during the Monte Carlo search, with the search terminating after 3 rounds. In the case of OPRO, one candidate prompt is generated per iteration, with the search halted after 15 iterations. For DPS, we generate prompts for 12 additional rounds, maintaining the same total number of candidate prompts as the other algorithms.

\subsection{Result and Analysis}
As demonstrated by Table \ref{tab:ado_performance}, employing ADO for data optimization consistently leads to comparable or superior performance across all datasets for all three LLM backbones, compared to inferencing with unoptimized data. Additionally, DPS outperforms both APE and OPRO in eight, seven, and nine out of nine datasets for GPT-3.5 Turbo, Gemini-1.5 Flash, and Llama-3.1 70B, respectively. This highlights the effectiveness of batch-based prompt search with diverse candidates. 

Furthermore, Table \ref{tab:ado_and_others_performance} demonstrates that integrating ADO with existing Prompt Engineering techniques, including CoT, ICL, and PE2, consistently results in a noticeable performance enhancement compared to employing these techniques alone across all nine datasets. For instance, ADO significantly enhances the effectiveness of CoT, particularly in the QA, Job, and FD datasets. For QA, applying CoT alone even results in slightly worse performance compared to not applying it, while combining CoT with ADO yields substantially better performance. These results demonstrate the complementarity of ADO towards both Instruction Optimization and Data Augmentation.

% This improvement is evident across all prompt search algorithms employed. Notably, CoT prompting amplifies the effectiveness of ADO, particularly on the QA, Job, and Financial Distress (FD) datasets, where it significantly outperforms non-CoT configurations. ten-shot ICL achieves the best results for e-commerce recommendation tasks, surpassing CoT prompting in these cases.

% In terms of prompt search algorithms, our proposed DPS consistently matches or outperforms both APE and OPRO across all datasets. The largest improvements are observed in the zero-shot CoT setting, where DPS shows the most significant gains, underscoring its effectiveness in optimizing prompts and enhancing overall model performance.

\begin{table*}[!ht]
\centering
\resizebox{14cm}{!}{
\begin{tabular}{l c c c c c c c c c c}
\toprule
Modeling variant & QA & Job & GSM & AB & AT & AE & CI & HD & FD & Mean\\ 
\hline
\hline
GPT & 0.578 & 0.619 & 0.285 & 0.124 & 0.129 & 0.211 & 0.788 & 0.617 & 0.639 & 0.443 \\ 
\hline
GPT w/ CoT & 0.571 & 0.663 & 0.698 & 0.127 & 0.137 & 0.198 & 0.827 & 0.678 & 0.688 & 0.510 \\
GPT w/ CoT + ADO & \textbf{0.679}& \textbf{0.807}& \textbf{0.851}& \textbf{0.185}& \textbf{0.219}& \textbf{0.257}& \textbf{0.879}& \textbf{0.751}& \textbf{0.789}& \textbf{0.602}\\ 
\hline
GPT w/ ICL &0.584 & 0.617 & 0.294 & 0.141 & 0.147 & 0.225 & 0.809 & 0.651 & 0.653 & 0.458 \\ 
GPT w/ ICL + ADO & \textbf{0.597} & \textbf{0.641}& \textbf{0.778}& \textbf{0.199}& \textbf{0.223}& \textbf{0.262} & \textbf{0.851} & \textbf{0.728}& \textbf{0.668}& \textbf{0.549}\\ 
\hline
GPT w/ PE2 & 0.592 & 0.634 & 0.301 & 0.162 & 0.152 & 0.209 & 0.838 & 0.649 & 0.685 & 0.469 \\
GPT w/ PE2 + ADO & \textbf{0.618} & \textbf{0.659} & \textbf{0.312} & \textbf{0.183} & \textbf{0.178} & \textbf{0.234} & \textbf{0.863} & \textbf{0.697} & \textbf{0.722} & \textbf{0.496} \\
\bottomrule
\end{tabular}
}
\vspace{-3pt}
\caption{Performance when ADO is combined with other Prompt Engineering techniques, using GPT-3.5 Turbo as the backbone (denoted as ``GPT''). ``CoT + ADO'' denotes applying both CoT and ADO, ``ICL + ADO'' denotes applying both ICL and ADO, and ``PE2 + ADO'' denotes applying both PE2 and ADO. For each dataset on each technique, any performance enhancement resulting from ADO integration is highlighted in bold.}
\label{tab:ado_and_others_performance}
\vspace{-9pt}
\end{table*}

\subsection{Ablation Study}
\label{sec:Ablation Study}
In this section, we perform an ablation study to assess the impact of different components of the ADO framework from three perspectives: (1) whether both content optimization and format optimization are necessary, (2) whether incorporating a factual-validation LLM (\text{LLM}$_{\mathcal{F}}$) improves performance, and (3) whether data-optimizing in-context examples yields performance gains. For experiment details, please refer to \ref{sec:ablation}. The results of all three experiments are presented in Table \ref{tab:ablation studies} in the Appendix. As the table demonstrates, both content and format optimizations are essential for performance: removing format optimization significantly reduced performance on recommendation datasets and the CI dataset, while removing content optimization led to declines on other datasets. Moreover, incorporating \text{LLM}$_{\mathcal{F}}$ for hallucination mitigation produced comparable or improved performance across all datasets, with the most significant gains on the QA, Job, and GSM datasets. Finally, optimizing in-context examples led to noticeable improvements, particularly on the Job, GSM, and FD datasets.

\section{Related Work}
\label{sec:related}
Numerous approaches have been proposed for modifying prompts to enhance LLM performance, such as In-Context Learning and Instruction Optimization. In-Context Learning concentrates on providing the LLM with additional in-prompt exemplars from the same task domain, typically in the form of input data paired with their corresponding labels or outputs \cite{wei2023larger, dong2022survey, shin2022effect}. This method capitalizes on the model's ability to generalize from in-prompt examples, enabling the LLM to better comprehend the expected output format and task-specific requirements based on the provided exemplars. 

Instruction Optimization aims to modify the instruction part of the prompt to improve LLM performance. For example, \citet{si2022prompting} points out that composing better instructions can greatly boost LLM's performance on task inferencing. \citet{wei2022chain} proposes CoT reasoning, which introduces immediate reasoning steps into the output generation process. As demonstrated by \cite{wei2022chain}, employing zero-shot CoT substantially improve LLM performance tasks including logical reasoning, fraud detection, among many others. Extending beyond manually crafted instructions, various studies have proposed automated methods to search for optimal instructions tailored to specific tasks \cite{zhou2023large, pryzant2023automatic, yang2024large}. For instance, APE \cite{zhou2023large} introduces an iterative Monte Carlo search to refine prompt instructions. It first uses an instruction-proposing LLM to generate a set of candidate instructions, then evaluates each on a validation set to select the best-performing candidates.

% This encourages the LLM to articulate its reasoning path, making the model's thought process more transparent and aligned with human expectations. 
Despite these advances, directly optimizing the presentation of input data has received little attention. In this work, we hypothesize that optimizing both the data content and format may yield performance improvement when employing LLM for task inferencing. Building on the principles of automatic prompt optimization, we propose a novel framework called Automatic Data Optimization (ADO). In ADO, an LLM, denoted as \text{LLM}$_{\mathcal{G}}$, iteratively proposes and searches data-optimization instructions aimed at maximizing LLM performance.

\section{Conclusions}
\label{sec:conclusion}
In this paper, we introduce a new sub-direction of prompt engineering: input data optimization, facilitated by the ADO framework and the DPS algorithm. The ADO framework automates content and format optimization by leveraging LLMs as universal domain experts, reducing the need for manual data processing. DPS enhances this process by generating diverse data optimization prompts, enabling broader exploration and increasing the likelihood of identifying optimal procedures. Empirical results demonstrate that ADO not only improves modeling performance when used alone but also further enhances performance when combined with other prompt engineering methods.

% Together, ADO and DPS offer an efficient, salable approach to data optimization, streamlining the preparation process and improving model performance. 

\section{Limitations}
As we explore the novel approach of input data optimization within prompts, we question whether it is possible to simultaneously search for both the optimal instruction and the optimal procedures for input data optimization in a specific inference task. Currently, as detailed in the paper, we first search for the optimal data representation using ADO, and then for the optimal instruction using PE2. However, this process involves two distinct steps, and it would be more efficient to search for both the instruction and data optimization concurrently. Therefore, in the future, we aim to investigate the feasibility of jointly optimizing both components, as proposed in \cite{sordoni2024joint, chen2024prompt}, to further enhance LLM performance.

\bibliography{custom}

\appendix

\section{Appendix}

\begin{table*}[t]
  \centering
  \footnotesize % This sets a smaller font size
  \begin{tabular}{l|c|c|c|c|c|c|c|c|c}
    \toprule
    & QA & Job & GSM & AB & AT & AE & CI & HD & FD \\
    \hline
    \hline
    ADO-Engineering & 0.667 & 0.789 & 0.843 & 0.155 & 0.177 & 0.229 & 0.839 & 0.742 & 0.776  \\
    ADO-Reformulation & 0.602 & 0.719 & 0.734 & 0.189 & 0.208 & 0.253 & 0.868 & 0.684 & 0.705 \\
    ADO w/ Factual-check & 0.691 & 0.823 & 0.864 & 0.187 & 0.221 & 0.262 & 0.884 & 0.747 & 0.795 \\
    ADO on ICL Samples & 0.599 & 0.682 & 0.803 & 0.187 & 0.228 & 0.267 & 0.871 & 0.734 & 0.691 \\
    \bottomrule
  \end{tabular}
  \caption{Ablation Study Performance.}
  \label{tab:ablation studies}
\end{table*}

\subsection{Bayesian Search Specifics}
\label{sec:bayes}

Bayesian Search is an informed search which achieves better performance than uninformed searches such as Random Search \cite{turner2021bayesian}. In this work, we propose to incorporate Bayesian Search as part of the data-optimization procedure search, by tuning $k$, $c_1$, and $c_2$ as ``hyper-parameters'' based on performance of the validation set $S$. This enables us to dynamically control both the number of candidate prompt to be generated per iteration for batch update, as well as the degree of diversity among candidate prompts.

\subsection{Ablation Study Specifics}

\label{sec:ablation}

\paragraph{Data Optimization Objectives}
We evaluate the effectiveness of the two optimization objectives—content optimization and format optimization—in ADO. To this end, we constrain the data-optimization prompt $\mathbf{P}_o$ to focus on either data engineering procedures (content optimization) or structural reformulation (format optimization), using zero-shot CoT as the prompting format. Specifically, we modify the meta-prompt $\mathbf{P}_m$ to explicitly prohibit instructions related to the non-evaluated aspect, ensuring $\mathbf{P}_o$is restricted to either content or format optimization. These are denoted as ``ADO-Engineering'' (data engineering only) and ``ADO-Reformulation'' (structural reformulation only).

\paragraph{Factual-validation LLM} We also investigate whether integrating the factual-validation LLM (\text{LLM}$_{\mathcal{F}}$), into the ADO workflow, as described in Section \ref{sec:imp} enhances performance. Using zero-shot CoT, we perform cross-validation on optimized input data, iterating between  \text{LLM}$_{\mathcal{F}}$ and  \text{LLM}$_{\mathcal{O}}$ until a consensus is reached or a maximum of four rounds is completed. If no consensus is reached, the optimized input from the final validation round is used for prompt construction. This configuration is referred to as ``ADO w/ Factual-check.''

\paragraph{Optimized Input for ICL} 
In Section \ref{sec:exp}, all in-context examples are presented in their unoptimized form. Here, we examine whether optimizing the input data of ICL examples, using the same procedures applied to the evaluation data, leads to improved performance. The hypothesis is that optimized in-context examples will better align with the evaluation input, facilitating easier learning for the LLM. We optimize the ICL input data and augment the prompt with these optimized examples paired with their respective outputs, denoted as ``ADO on ICL Samples.''

Table \ref{tab:ablation studies} presents the ablation study results. For the first experiment: both data engineering and structural reformulation are crucial for maintaining performance. Limiting optimization to data engineering led to a significant drop in performance on all recommendation datasets and the CI dataset, while restricting optimization to structural reformulation resulted in performance degradation on the other datasets. For the second experiment: incorporating \text{LLM}$_{\mathcal{F}}$ for factual validation produced comparable or improved performance across all datasets, with the most significant gains on the QA, Job, and GSM datasets. Finally, optimizing in-context examples led to noticeable improvements, particularly on the Job, GSM, and FD datasets.

\vspace{10pt}
\begin{lstlisting}[language=HTML, caption=Meta Prompt Example]
Dataset Description: <description>

Your task is to propose a creative,
detailed, and step-by-step algorithm
to enrich and then reformulate samples
in this dataset. The goal of the
algorithm is to perform thorough
data engineering and reformulation on
the sample, so that it is easier for
an LLM to generate the target outputs.

Below are some example dataset samples
with target outputs as references:

Examples:
- <sample input1>; Output: <sample output1>
- <sample input2>; Output: <sample output2>
- <sample input3>; Output: <sample output3>
- ...

Please Note:
- Do NOT refer to any external database.
- Do NOT perform vector generations.
- ONLY propose steps that an LLM
  can do on its own.
- ...

Below is a list of prior-proposed data 
optimization algorithms, provided to 
you as additional context:
- Algorithm 1; Score: a1
- Algorithm 1; Score: a2
- ...

\end{lstlisting}
\vspace{-5pt}

\label{sec:appendix}

% \subsection{If needed} 
% \label{if needed}

\end{document}